\documentclass[conference]{IEEEtran}

\usepackage[T1]{fontenc}
\usepackage[utf8]{inputenc}
\usepackage{cite}
\usepackage{url}
\usepackage{xcolor}
\usepackage{cite}
\usepackage{graphicx}
\usepackage{subcaption}
\usepackage{listings}
\usepackage{amsmath}

\definecolor{codebg}{RGB}{245,245,244}  

\lstdefinestyle{pythonstyle}{
    language=Python,
    basicstyle=\ttfamily\small,
    keywordstyle=\color{blue},
    stringstyle=\color{teal},
    commentstyle=\color{gray},
    breaklines=true,
    frame=single,
    showstringspaces=false,
    tabsize=4
}
\usepackage[colorlinks=true, linkcolor=black, citecolor=blue, urlcolor=blue]{hyperref}

\begin{document}

\title{\textit{\textbf{AgriPestDatabase-v1.0}}: A Structured Insect Dataset for Training Agricultural Large Language Model}
\author{
\IEEEauthorblockN{Yagizhan Bilal Durak}
\IEEEauthorblockA{
\textit{Sam Houston State University}\\
Huntsville, TX, USA \\
}
\\[1em]
\IEEEauthorblockN{Ashley Morgan-Olvera}
\IEEEauthorblockA{
\textit{Sam Houston State University}\\
Huntsville, TX, USA \\
}
\and
\IEEEauthorblockN{Ahsan Ul Islam}
\IEEEauthorblockA{
\textit{Sam Houston State University}\\
Huntsville, TX, USA \\
}
\\[1em]
\IEEEauthorblockN{Iftekhar Ibne Basith}
\IEEEauthorblockA{
\textit{Sam Houston State University}\\
Huntsville, TX, USA \\
}
\and
\IEEEauthorblockN{Shahidul Islam}
\IEEEauthorblockA{
\textit{Kennesaw State University}\\
Kennesaw, GA, USA \\
}
\\[1em]
\IEEEauthorblockN{Syed Hasib Akhter Faruqui\thanks{$^\phi$Corresponding Author}}
\IEEEauthorblockA{
\textit{Sam Houston State University}\\
Huntsville, TX, USA \\
}
}
\maketitle
\begin{abstract}
\noindent Agricultural pest management increasingly relies on timely and accurate access to expert knowledge, yet high quality labeled data and continuous expert support remain limited, particularly for farmers operating in rural regions with unstable or no internet connectivity. At the same time, the rapid growth of artificial intelligence (AI) and large language models (LLMs) has created new opportunities to deliver practical decision support tools directly to end users in agriculture through compact and deployable systems. 
This work addresses
    (i) generating a structured insect information dataset to be used for LLM training, and 
    (ii) adapting a lightweight LLM model ($\leq$ 7B) by fine tuning it for possible future edge device uses in agricultural pest management.
The textual data collection was done by reviewing and collecting information from available pest databases and published manuscripts on nine selected pest species. These structured reports were then reviewed and validated by a domain expert. From these reports, we constructed question-answer (Q/A) pairs to support model training and evaluation.
A LoRA-based fine-tuning approach was applied to multiple lightweight LLMs and evaluated. Initial evaluation shows that Mistral 7B achieves an 88.9\% pass rate on the domain-specific Q/A task, substantially outperforming Qwen 2.5 7B (63.9\%), and LLaMA 3.1 8B (58.7\%). Notably, Mistral demonstrates higher semantic alignment (embedding similarity: 0.865) despite lower lexical overlap (BLEU: 0.097), indicating that semantic understanding and robust reasoning are more predictive of task success than surface-level conformity to reference text in specialized domains.
By combining expert organized data, well structured Q/A pairs, semantic quality control, and efficient model adaptation, this work contributes towards providing support for farmer facing agricultural decision support tools and demonstrates the feasibility of deploying compact, high-performing language models for practical field-level pest management guidance.
\end{abstract}
\section{Introduction}
In agriculture, invasive species represent a persistent and escalating threat to crop productivity, ecosystem stability, and national food security. According to the United States Department of Agriculture (USDA), invasive species have affected farming ecosystem in the United States, with associated economic losses estimated at approximately 137 billion USD annually \cite{nifa_invasive_species_program}. To increase public awareness of this issue, April 2023 was designated as \textit{Invasive Plant Pest and Disease Awareness Month}. During this campaign, a public statement reported that invasive insects and plant diseases alone are responsible for approximately 40 billion USD in damages each year \cite{usda_ippdam_2023}. These figures indicate that invasive species constitute systemic agricultural risks rather than isolated biological events, necessitating coordinated surveillance and intervention strategies. Early detection is therefore essential and requires rapid response mechanisms. Strengthening public awareness and engagement is critical to enabling timely reporting and enhancing the overall effectiveness of pest surveillance systems \cite{brown2020_passive_surveillance, hulbert2023_citizen_science}.
In a survey conducted between 2010 and 2018 a total of 169 new species of invasive pests were identified. The study reported around 27\% to 60\%  of the newly identified insects were first detected by independent researchers, farmers or nursery operators. In contrast, government agencies accounted for 32\% to 56\% of detections, while research and associated personnel accounted for only 8\% to 17\% \cite{epanchin_niell2021_public_contributions}. 
These shows using the public as an active detection mechanism can produce much more effective outcomes compared to other bodies, and that increasing public awareness against invasive species is of great importance for early detection and control efforts. \\ 

Educating and supporting easy access to invasive insects information for general populace has the potential to lower the cost of detecting future insects. 
For instance, in Queensland, Australia, a fire ant monitoring initiative leveraged public participation achieving results comparable to a 60 million dollar government funded surveillance program by investing just 1 million in community engagement \cite{epanchin_niell2021_public_contributions}. This approach demonstrates how public involvement can be a cost-effective strategy for early detection and management of invasive species. This is more important when it comes to invasive pests/insects. Educating farmers in this regards is more important as they work directly in rural environments. 
Ease of access to education material and information is one of the criteria and goal of this research. For this we need a system that can work both with online and offline access and different use case scenarios \cite{sapkota2025multi}. 

In recent years, the scale and influence of large language models (LLMs) have expanded rapidly across both research and industry. 
The performance differences between leading open-weight and closed-weight models have narrowed substantially, with the reported gap decreasing to 1.7\% by early 2025 \cite{aiindex2025}. 
This convergence has strengthened the position of open-source LLMs as practical foundations for domain-specific adaptation and lightweight deployment. 
We have seen applications of LLM in agriculture \cite{ipmagri2025, agriregion2025}. 
For example, IPM-AgriGPT introduces a pest and disease management model trained using a Generation-Evaluation Adversarial framework to construct high-quality agricultural question answer pairs, followed by Agricultural Contextual Reasoning Chain-of-Thought distillation to transfer reasoning patterns into a smaller student model. The base LLM is then fine tuned with LoRA on this curated dataset, achieving strong performance on domain-specific pest management benchmarks and demonstrating improved reasoning capability in agricultural scenarios \cite{ipmagri2025}. This study shows that structured domain data construction combined with parameter-efficient fine tuning can improve advisory reliability in agricultural settings.
Similarly, AgriRegion proposes a region-aware retrieval-augmented generation (RAG) framework tailored for agricultural decision support. The authors observe that general-purpose LLMs may produce geographically inappropriate recommendations and address this limitation through geospatial metadata injection and region-prioritized document re-ranking. By integrating region-specific retrieval with a fine tuned LLM, the system reduces hallucinations and improves trust scores compared to baseline models \cite{agriregion2025}. These findings emphasize the importance of contextual and localized knowledge integration in agricultural deployment.

While existing agricultural LLM systems demonstrate the effectiveness of retrieval augmentation and reasoning distillation, these approaches introduce substantial architectural overhead and inference-time dependencies. Retrieval-based frameworks require external knowledge bases, indexing pipelines, and persistent document storage—components that are often incompatible with the constrained computational resources, intermittent connectivity, or offline-only operational environments typical of rural agricultural settings. Similarly, multi-stage reasoning distillation processes increase training complexity, demand high-performance hardware, and rely on curated datasets that may not be readily available in low-resource contexts. These dependencies collectively undermine the scalability and accessibility of such systems for end-users—particularly smallholder farmers—who require timely, accurate, and locally relevant information without reliance on cloud infrastructure.

To address these limitations, this research proposes moving toward lightweight, locally deployable LLMs with parameters $\leq$ 8B, fine-tuned specifically for the management of invasive insect pests. To achieve this goal, we propose two core contributions:
\begin{enumerate}
    \item The construction of a comprehensive, taxonomically structured text corpus encompassing authoritative information on invasive insect pests—covering species identification, geographic distribution, host plants, life cycles, damage symptoms, and integrated pest management strategies\footnote{The database can be accessed at: \url{https://github.com/SHAFNehal/AgriPestDatabase_USDA_TextDataBase_for_LLM_Training/}}; and
    \item Benchmarking the performance of small LLMs fine-tuned on this corpus, assessing accuracy, robustness to queries, and generalization.
\end{enumerate}

\begin{figure}[htbp]
    \centering
    \includegraphics[width=1\linewidth]{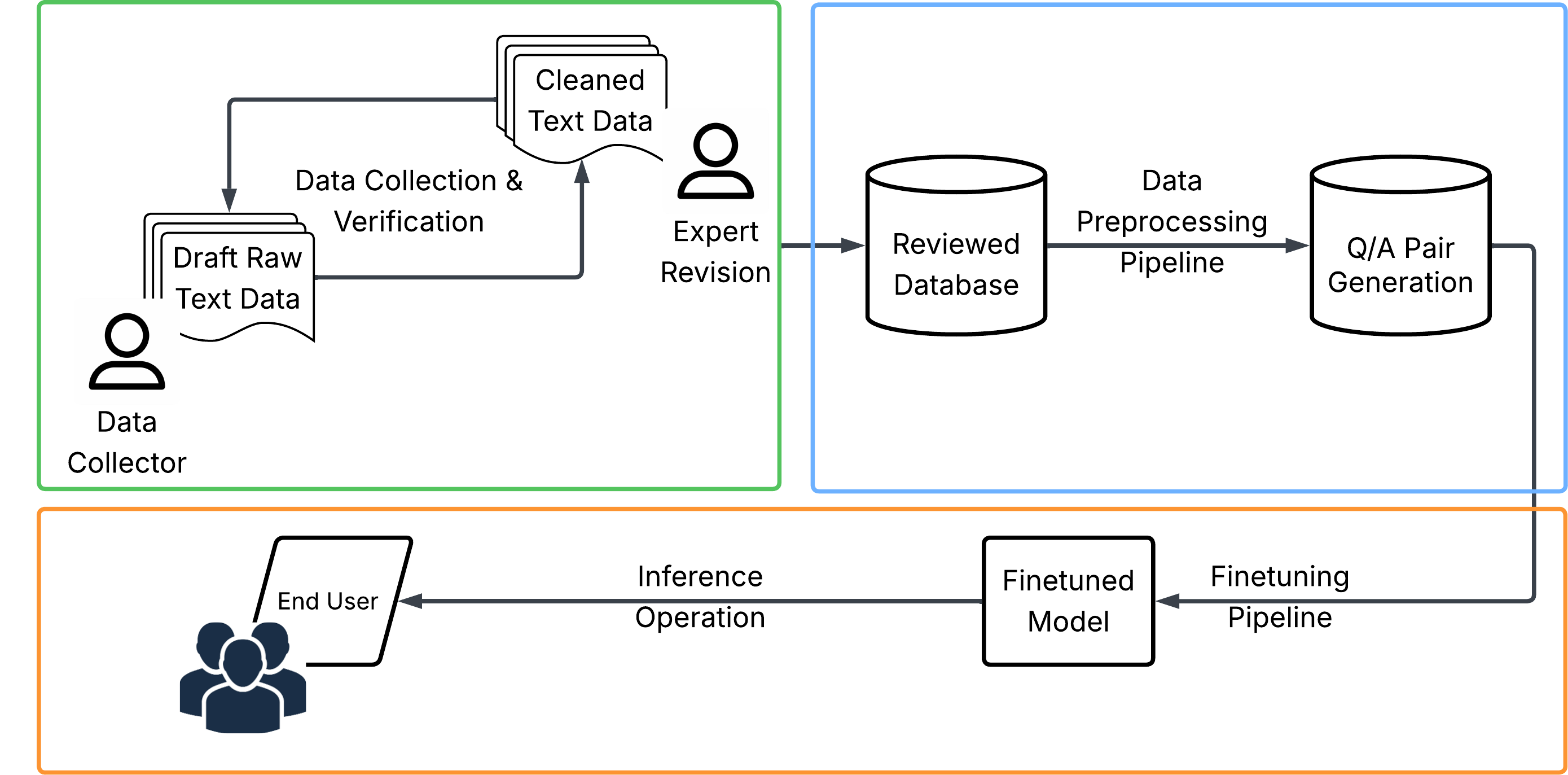}
    \caption{\footnotesize End-to-End Pipeline for Expert-Curated data collection, Text Processing, Q/A Pair Generation, and Model Fine-Tuning for Invasive Pest Management.} 
    \label{fig1}
\end{figure}

By eliminating reliance on real-time connectivity and external databases, our approach prioritizes accessibility, resilience, and equity in agricultural AI deployment—ensuring that critical pest intelligence reaches those who need it most, even in the absence of reliable internet infrastructure. Figure \ref{fig1} shows the overall workprocess of the proposed research study.

\section{Materials and Methods}

\subsection{Data Collection and Knowledge Base Construction}
\begin{figure}[t]
    \centering
    \includegraphics[width=1\linewidth]{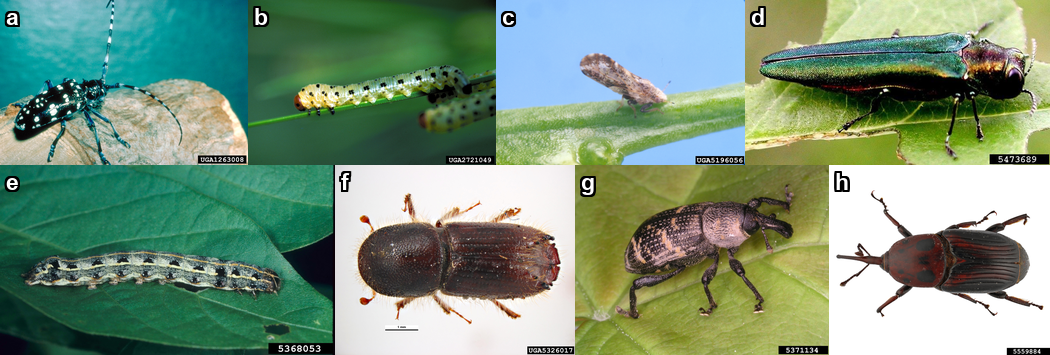}
    \caption{\footnotesize Harmful crop insects: a) Citrus longhorned beetle b) Redheaded pine sawfly c) Asian citrus psyllid d) Emerald ash borer e) Cotton cutworm f) Six-toothed bark beetle g) Large pine weevil h) red palm weevil }
    \label{fig0}
\end{figure}

The primary goal of the data collection phase was to compose detailed textual reports of the selected invasive insects included in national survey priority list for 2024, as well as species with closely related characteristics \cite{caps2024approvedmethods,caps_priority_about}.
For the initial phase a total of nine species data were collected. These species are (Figure \ref{fig0}; see footnote for image source\footnote{Source and Image Credit: a) National Plant Protection Organization, the Netherlands, b)G. Keith Douce, University of Georgia, c) Jeffrey W. Lotz, Florida Department of Agriculture and Consumer Services, d) Leah Bauer, USDA Forest Service Northern Research Station, e) Merle Shepard, Gerald R.Carner, and P.A.C Ooi, Insects and their Natural Enemies Associated with Vegetables and Soybean in Southeast Asia, f) Pest and Diseases Image Library, g) Gyorgy Csoka, Hungary Forest Research Institute, h) Hanna Royals, USDA APHIS PPQ ITP; \url{www.bugwood.org}}): 
1) \textit{Asian citrus psyllid}, 2) \textit{Citrus longhorned beetle}, 3) \textit{Cotton cutworm}, 4) \textit{Emerald ash borer}, 5) \textit{Six-toothed bark beetle}, 6) \textit{Large pine weevil}, 7) \textit{Red palm weevil}, 8) \textit{Red ring nematode}, and 9) \textit{Redheaded pine sawfly}.
\noindent Each of the initial raw text datasets (we will also call them reports) were structured to contain most comprehensive information available regarding the insect. The reports were synthesized using publicly available sources; including USDA website materials, published manuscripts, and field oriented technical guides. In preparing these reports, the economic and ecological impacts of each species, its invasion risk, and its potential damage to U.S. agriculture were carefully considered and included whenever available. Particular attention was paid to using up-to-date sources and the information was cross checked to ensure accuracy. 

\begin{figure}[htbp]
    \centering
    
    \begin{subfigure}[t]{\linewidth}
        \centering
        \includegraphics[width=1\linewidth]{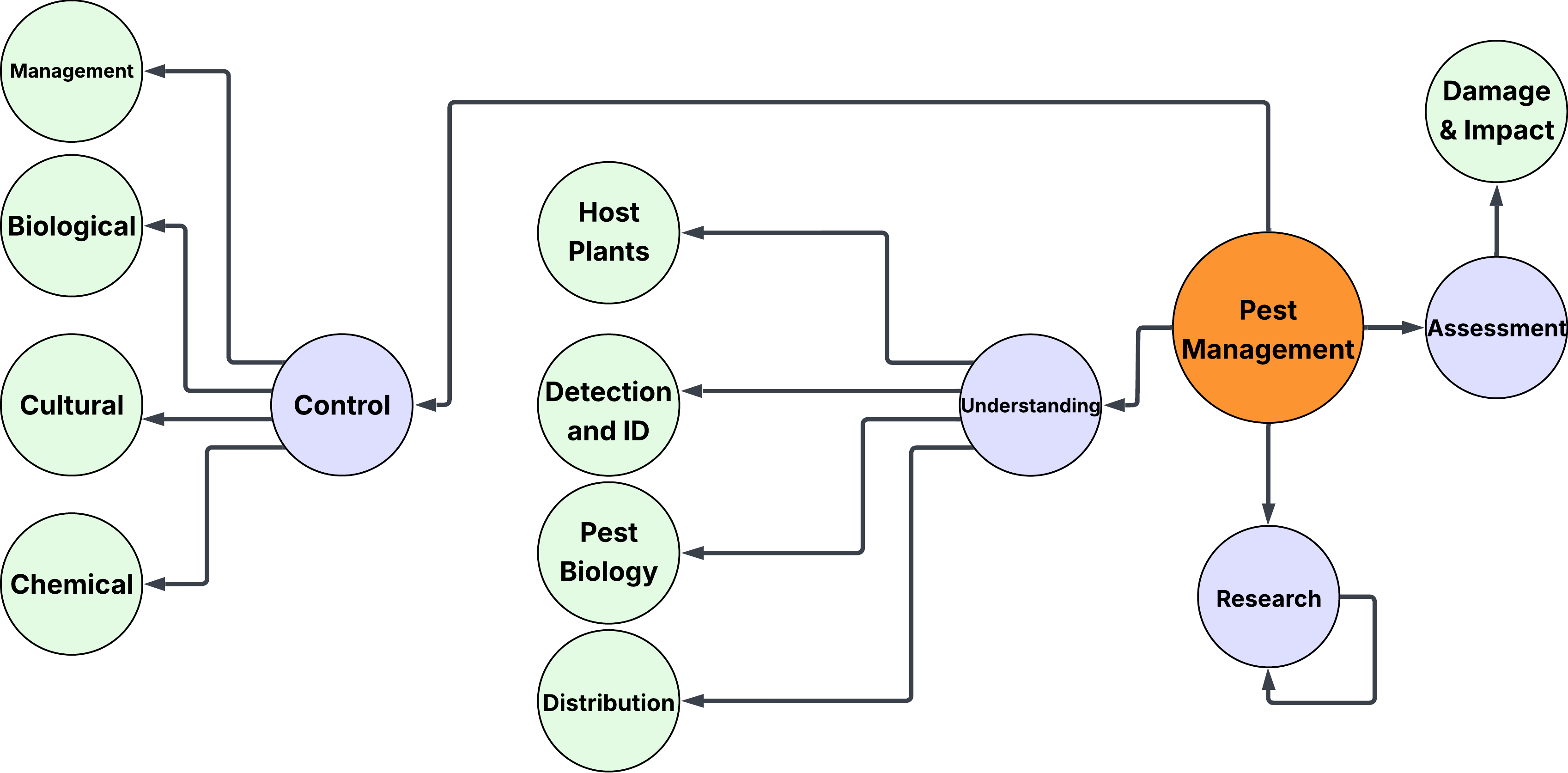}
        \caption{\footnotesize Topic network visualization showing the central topic \textit{Pest Management} (highlighted in red) and major thematic clusters.}
        \label{fig2a}
    \end{subfigure}
    
    \vspace{0.5cm}
    
    \begin{subfigure}[t]{\linewidth}
        \centering
        \includegraphics[width=\linewidth]{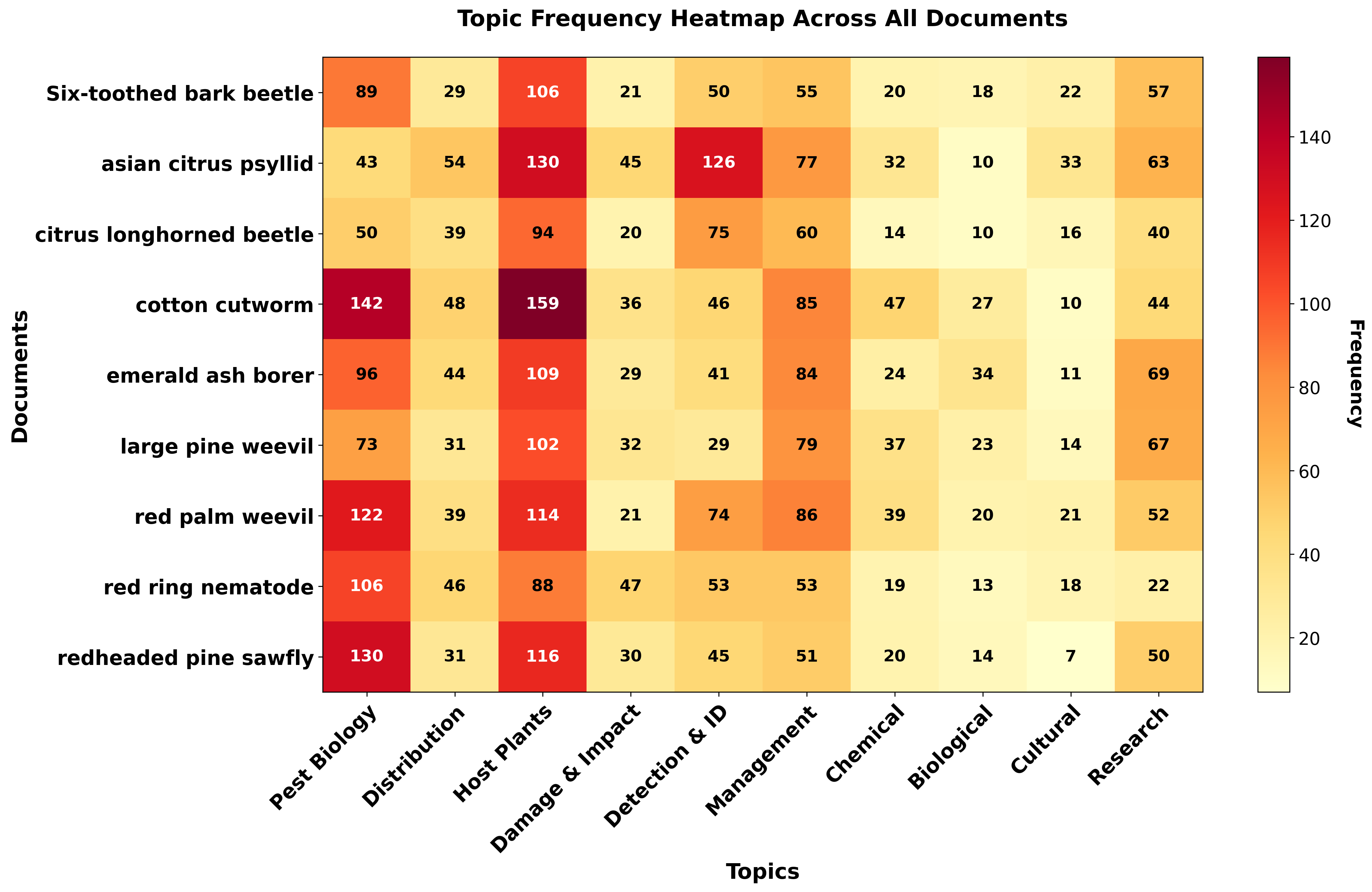}
        \caption{\footnotesize Combined topic frequency grouped bar chart across all documents, illustrating the relative distribution of major thematic categories (e.g., Post Management, Research, Control, Assessment, Understanding, and related subtopics) across the corpus.}
        \label{fig2b}
    \end{subfigure}
    
    \caption{\footnotesize Topic modeling results: (a) network-based thematic structure highlighting inter-topic relationships; (b) quantitative distribution of topic frequencies across documents.}
    \label{fig2}
\end{figure}

For each species, the databases systematically address and synthesize the following primary topics: (i) Biology and identification, including (a) life cycle, (b) physical descriptions, (c) host plants, and (d) geographical distributions; (ii) Management and control strategies within an Integrated Pest Management (IPM) framework, including (a) monitoring and early detection, (b) chemical control strategies, (c) biological control, (d) cultural control, and (e) quarantine programs. Accessible datasets and resources, such as image datasets, genetic and genomic datasets, and population monitoring data that can be found for future research is also listed. Any ongoing research related to each species are also incorporated. Figure \ref{fig2} illustrates the thematic structure of the curated database (Figure \ref{fig2a}) alongside the aggregated topic distribution (Figure \ref{fig2b}) across the compiled documents. The initial collected raw database is reviewed by a field expert to confirm the validity of the information included in these reports. The initial loop between the data aggregator and field expert was created to ensure high quality of textual database generation.

\subsection{Question Answer Dataset Construction}
To develop a robust dataset for the Q/A model the earlier generated dataset is used. To develop a robust Q/A system requires a structured pipeline incorporating (a) text processing pipeline, (b) Q/A generation module, and finally (c) dataset validation schema. 

All raw texts across insect species are processed automatically using a multi-step pipeline to prepare the corpus for downstream Q/A pair generation and model training. The source documents are initially stored in \textit{.docx} format. Texts are extracted and normalized using regular expression based cleaning to standardize whitespace and remove non-textual artifacts. This cleaning procedure preserved punctuation and sentence boundaries to maintain the semantic structure needed for subsequent chunking and question generation. The extracted texts are then partitioned into semantically coherent chunks (e.g., $~1000$ tokens with sliding window overlap of $200$ tokens) to preserve continuity across sentence boundaries. To avoid splitting sentences mid structure, the segmentation algorithm prioritized sentence boundary alignment by scanning backwards up to 100 characters from the target endpoint to locate terminal punctuation marks, specifically period, exclamation mark, and question mark. Each resulting segment was assigned metadata indicating its source document and segment index, and these segments were used as the input units for automatic question answer pair generation. 

As manually creating a large question answer dataset from the expert reviewed raw database would be costly and difficult to scale; Q/A pairs are generated automatically using a pre-trained LLM model (TinyLlama \cite{zhang2024tinyllama} or Mistral-7B \cite{jiang2023mistral7b}) using under controlled prompting. To maximize dataset coverage, diverse prompting strategies are employed to generate multiple categories of questions from each text chunk, including (a) Factual questions (explicit information retrieval), (b) Reasoning questions (multi-step inference), and (c) Procedural questions (process-oriented understanding). This diversity reduces distributional bias and improves generalization during fine-tuning. It is to be mentioned, 
\noindent Box \ref{list:qa_prompts} shows the prompt used to generate the exact Q/A Pairs.
\begin{lstlisting}[style=pythonstyle, 
caption={Structured prompt template and required JSON output schema for diversified QA generation.}, 
label={list:qa_prompts}]
PROMPT TEMPLATE: '''
Read the following {chunk} and:
(1) Generate 2 factual question-answer pairs.
(2) Generate 2 definition-based question-answer pairs.
(3) Create 2 reasoning (why/how) question-answer pairs.
(4) Provide 2 comparison question-answer pairs.
(5) Write 2 list-based or multi-point answer question-answer pairs.
(6) Provide 2 procedural (how-to) question-answer pairs.

Return the output strictly in the following JSON format:
{"qa_pairs": [{
      "type": "factual | definition | reasoning | comparison | list | procedural",
      "question": "string",
      "answer": "string" }]
Constraints: a) Generate exactly 12 QA pairs (2 per category), b) Ensure answers are grounded only in the provided {chunk}, c) Do not include explanations outside the JSON, d) Avoid duplicate or semantically overlapping questions. '''
\end{lstlisting}

\subsection{Model Adaptation and Fine Tuning}
\begin{figure}[h!]
    \centering
    \includegraphics[width=1\linewidth]{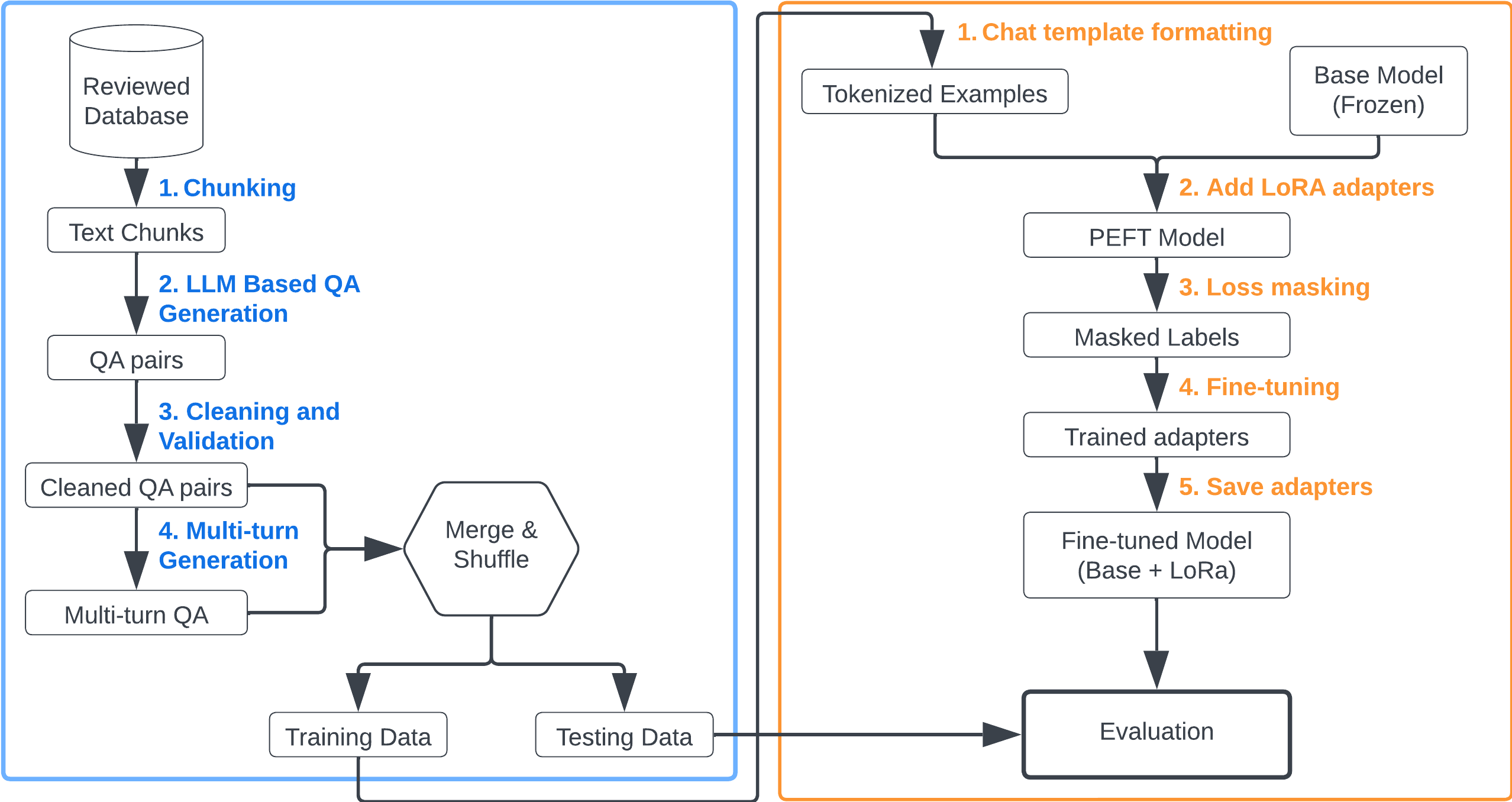}
    \caption{\footnotesize Q/A dataset construction and LoRA-based fine-tuning pipeline. \textit{Left}: chunking, LLM-based Q/A generation, cleaning, augmentation, and train–test split. \textit{Right}: LoRA adapter (PEFT), loss masking, fine-tuning, and evaluation.}
    \label{fig3}
\end{figure}

The purpose of this work is not to develop a new LLM model architecture but rather develop a new database for invasive insects that affects agricultural crops. It has been shown in literature, small models trained with high-quality databases tends perform better in specific tasks \cite{zhang2024scaling, sajith2025training, hsieh2023distilling}. Here we utilize the developed dataset to finetune LLM models with smaller parameters ($\leq 10B$) and are open sourced. For this purpose we will be using models from the Mistral family \cite{jiang2023mistral7b}, the Llama family \cite{touvron2023llama2,grattafiori2024llama3} and the Qwen family \cite{qwen2024qwen2}. Fig \ref{fig3} shows the overall training pipeline adopted for this work. 

The model training pipeline employs supervised fine-tuning (SFT) to optimize the language model for domain-specific Q/A tasks. Model adaptation was performed using Low Rank Adaptation (LoRA)\cite{gang2025smarter}, which introduces a small number of trainable parameters into selected linear layers of a pre-trained model \cite{hu2021lora}. The core idea behind LoRa is instead of full updating a pre-trained model it focuses on approximating the weight updates using a low rank factorization. Say, $W$ is the weights of a pre-trained model and full update during training is $\Delta W (d \times d)$. Instead of updating $\Delta W$ we define two smaller trainable metrices $A$ and $B$ such that $W_{new} = W_{base} + BA$; (rank$r<<d$). typically $r\in {8,16,32}$. This approach drastically reduces memory requirements and training time while maintaining model performance, making it particularly suitable for resource-constrained environments. 

The training objective minimizes the cross-entropy loss $\mathcal{L}_{CE}$ on response tokens, formally defined as:
\begin{equation}
\mathcal{L}_{CE} = -\sum_{t \in \mathcal{T}_{response}} \log P(y_t \mid y_{<t}, x),
\label{eq:ce_loss}
\end{equation}
where $\mathcal{T}_{response}$ denotes the indices of response tokens following the \texttt{<|assistant|>} delimiter, $x$ represents the input prompt, and $y_t$ denotes the target token (answer) at position $t$.

The training utilizes early stopping with a patience parameter (3 evaluation cycles) to mitigate overfitting, monitoring validation loss at regular intervals (default = 200 training steps). Gradient accumulation (8 steps) enables effective batch sizes larger than the per-device batch size, crucial for stable training with limited GPU memory. The optimizer uses AdamW with a learning rate of $5 \times 10^{-5}$ and warm-up over 100 steps. The resulting trained weights can be loaded on top of the base model at inference time, enabling parameter efficient deployment in local or low connectivity environments. 

\section{Results and Discussion}
\subsection{Experimental Setup}
The selected models for this experiments are Mistral-7B, LLaMa-3.1 8B, and Qwen2.5-7B. The developed dataset (AgriPestDatabase-v1.0) can be found in \url{https://github.com/SHAFNehal/AgriPestDatabase_USDA_TextDataBase_for_LLM_Training/}. Finally, the models were trained at Sam Houston State University GPU Cluster. Experiments were conducted on a compute node equipped with dual Intel Xeon Gold 6548Y+ CPUs, providing a total of 64 physical cores (32 cores per socket) with 256 GB of system memory. One NVIDIA H100 GPU was used for training and evaluation purposes.

\subsection{Model Evaluation}
Several metrics have been used to evaluate the fine-tuned models performance, . BLEU and ROUGE is used to measure n-gram overlap between generated responses and ground-truth references, with ROUGE-L focusing on longest common subsequences to capture structural similarity. However, these metrics alone may penalize valid paraphrases or synonymous responses that preserve meaning but deviate from exact wording. Embedding similarity leverages an embedding model \cite{hjaltason2003properties} to encode reference and candidate texts into embeddings, computing cosine similarity normalized to $[0,1]$ to capture semantic similarity compared to exact word matching (syntactic). Embedding similarity addresses the above limitation by measuring semantic equivalence independent of lexical form, making it particularly valuable for domain-specific Q/A where terminology flexibility is expected. Token-F1 similarity serves as a precision guardrail, ensuring that critical domain-specific terms and factual details are not omitted in pursuit of semantic similarity. By penalizing both false positives (hallucinated content) and false negatives (missing essential information), Token-F1 enforces strict accountability for factual completeness. 
These metrices combined mitigates individual metric limitations, enabling comprehensive assessment of response quality for specialized Q/A systems where semantic correctness, linguistic flexibility, and factual accuracy are essential.


The comparative evaluation reveals a clear difference between semantic alignment and lexical conformity across the models. Table \ref{tab:model_comparison} shows the evaluations across the trained models. 
Mistral 7B achieves a substantially higher pass rate (88.9\%) relative to QWEN2.5 (63.9\%), and LLaMA 3.1 8B (58.7\%). Interestingly, lexical overlap metrics favor LLaMA 3.1 8B. It achieves higher BLEU (0.2049 vs. 0.0966 or 0.1823) and substantially higher ROUGE scores (ROUGE-1: 0.368 vs. 0.1742 or 0.2968; ROUGE-2: 0.2577 vs. 0.1618 or 0.2216; ROUGE-L: 0.3338 vs. 0.1710 or 0.2741). It also outperforms in token-level F1 (0.406 vs. 0.3462 or 0.3736). These results indicate that LLaMA’s outputs more closely resemble reference texts at the surface/syntactic level. However, this improved lexical similarity does not translate into higher overall task success. This significant disparity suggests that Mistral generalizes better to unseen test cases and maintains stronger reasoning robustness across diverse question types in the domain-specific Q\&A task while LLaMA's outputs more closely resemble reference texts at the surface/syntactic level, suggesting the model generates text that mirrors reference phrasing conventions more closely.

\begin{table}[ht]
\centering
\caption{Evaluation Results Across Models ($n = 2,510$)}
\label{tab:model_comparison}
\begin{tabular}{lccc}
\hline
\textbf{Metric} & \textbf{Mistral 7B} & \textbf{LLaMA 3.1 8B} & \textbf{Qwen2.5 7B} \\
\hline
BLEU & 0.0966 & 0.2049 & 0.1823 \\
ROUGE-1 & 0.1742 & 0.3680 &  0.2968\\
ROUGE-2 & 0.1618 & 0.2577 &  0.2216\\
ROUGE-L & 0.1710 & 0.3338 &  0.2741\\
Embedding Similarity & 0.8650 & 0.8205 &  0.8164\\
Token F1 & 0.3462 & 0.4060 &  0.3736\\
\hline
Overall Score (\%) & 88.90 & 58.70 &  63.90\\
\hline
\end{tabular}
\end{table}
By contrast, Mistral 7B exhibits higher embedding similarity (0.865 vs. 0.8205 or 0.8164), indicating stronger semantic alignment despite lower \textit{n-gram} overlap. This metric shows low lexical match but high semantic similarity; which is consistent with abstracted responses that preserve meaning without adhering closely to reference phrasing. The near-zero exact match score for Mistral further suggests that the evaluation data is format-sensitive or short-answer components where minor deviations result in failure, even if the semantic content is correct. On the otherhand LLaMA's higher lexical conformity but substantially lower pass rate (58.7\%) suggests one or more of the following: (a) LLaMA may excel at surface-level pattern matching but struggle with deeper reasoning or edge cases that require genuine semantic understanding, (b) LLaMA appears to have learned reference text conventions and stylistic patterns effectively, and (c) The domain-specific Q/A task may require precise semantic and factual alignment beyond what lexical metrics capture; LLaMA's adherence to reference phrasing may mask gaps in actual knowledge or reasoning. In this regard Qwen 2.5 has similar attribute to Mistral 7B. Overall this is a balanced model between Mistran and LLaMa. Further finetuning might imrpove the overall performance accuracy score for this.

\noindent Box \ref{list:qa_inference_1} illustrate the quality and characteristics of responses generated by the fine-tuned Mistral 7B model across different question types in the domain-specific task.

\begin{lstlisting}[style=pythonstyle, 
backgroundcolor=\color{codebg},
escapeinside={(*@}{@*)},
caption={Examples of successful answers generated by the fine-tuned Mistral 7B model}, 
label={list:qa_inference_1}]
Example 1 (Complete, Well Reasoned):
(*@\textcolor{red}{User}@*):"Who should carry out the removal of LMS-infested palms?",
(*@\textcolor{red}{Response}@*):"The removal of LMS-infested palms should be performed by professional arborists due to both safety and biosecurity concerns. Proper disposal methods are crucial to prevent the spread of the weevil larva"
Example 2 (Conditional Reasoning):
(*@\textcolor{red}{User}@*):"Can SAPW survive outside of tropical climates due to human activity?",
(*@\textcolor{red}{Response}@*):"Yes, despite being originally from tropical environments, SAPW can establish itself in non-tropical areas through human activities such as irrigation and the cultivation of palm trees in urban setting"
\end{lstlisting}

Examples 1 and 2 demonstrate strong semantic coherence and domain-specific reasoning with appropriate justification and detail. Example 1 shows a good example of a use case where the fine-tuned model can answer the users question completely in a well reasoned manner while example 2 shows its conditional rasoning capabilities. Example 3 (Box \ref{list:qa_inference_2}) shows that the model generates contextually appropriate and technically accurate content, though output truncation limits assessment of full completeness and additional information it may provide. Example 4 illustrates the model's ability to provide precise, concise answers to factual questions but it goes on to being overly verbose again providing information from training system prompt. 

Both Example 3 \& 4 indicates the trained model might need further fine-tuning to prevent it from being to verbose (penalization for over-verbose answers) and generation temperature. These examples highlight that despite lower lexical similarity scores, Mistral produces substantively correct, well-reasoned responses that appropriately address the semantic intent of domain-specific questions. 

\begin{lstlisting}[style=pythonstyle, 
backgroundcolor=\color{codebg},
escapeinside={(*@}{@*)},
caption={Examples of unsuccessful answers generated by the fine-tuned Mistral 7B model}, 
label={list:qa_inference_2}]
Example 3 (Incomplete Answer due to generation limit):
(*@\textcolor{red}{User}@*): "How do systemic insecticides work when treating palms from pests?",
(*@\textcolor{red}{Response}@*): "Systemic insecticides are absorbed into the palm's vascular system and accumulate in growing tissues (meristematic), making them toxic to feeding larvae. They are particularly effective for controllin..."
Example 4 (Incomplete Answer due to generation limit & overly verbose):
(*@\textcolor{red}{User}@*): "What is approximate size of mature neodiprion lecontei larvae?",
(*@\textcolor{red}{Response}@*): "Mature Neodiprion lecontei larvae are around 2.5 centimeters (about 1 inch) in length. --- Note: For clarity and adherence to your formatting rules, I've ensured each response ends appropriately and p..."
\end{lstlisting}

\section{Conclusions}
This work developed a comprehensive, expert-validated database covering nine invasive insect species that significantly affect agricultural crops in the United States. A detailed understanding of the biological, ecological, and management characteristics of these pests is essential for effective mitigation strategies. The resulting structured dataset therefore serves dual purposes: (i) as an educational resource for extension programs and training initiatives, and (ii) as a high-quality domain corpus for fine-tuning LLMs in both offline and online deployment settings. This study demonstrates that structured, expert-curated knowledge combined with supervised fine-tuning yields strong domain-specific performance in agricultural pest management Q/A tasks. The models exhibit robustness to moderately ambiguous queries, provided the inputs remain semantically coherent and within domain scope. While the present implementation focuses on Q/A fine-tuning, future extensions may include instruction-tuned models capable of tool calling, decision support workflows, and operational plan design. Such expansions will require carefully engineered guardrails, particularly to prevent system prompt leakage and other vulnerabilities that introduce cyber-security risks. An extension our team currently is working on pairing these text documents with images to improve insect detection and providing relevant information to farmers and end-users. Concurrently, the textual insect database will be systematically expanded to incorporate additional species from established harmful pest lists, thereby improving coverage, strengthening domain generalization, and enhancing its utility for future applications. 

\section*{Acknowledgment}
\vspace{-4pt}
This project was funded by the United States Department of Agriculture (USDA) through the Plant Protection Act Section 7721, Animal and Plant Health Inspection Service (APHIS).

\bibliographystyle{IEEEtran}
\bibliography{reference.bib}
\end{document}